\documentclass{article}

\usepackage{microtype}
\usepackage{graphicx}
\usepackage{subfigure}
\usepackage{booktabs} 

\usepackage{hyperref}

\usepackage[accepted]{icml2021}

\usepackage[utf8]{inputenc} 
\usepackage[T1]{fontenc}  
\usepackage{url}      
\usepackage{booktabs}    
\usepackage{amsfonts}    
\usepackage{nicefrac}    
\usepackage{microtype}   
\usepackage{shortcuts}
\usepackage{collcell}
\usepackage{amsthm}
\usepackage{amsmath}
\usepackage{amssymb}
\theoremstyle{plain}

\newtheorem{proposition}{Proposition}

\usepackage{wrapfig}
\usepackage{url}
\usepackage{color, colortbl}
\definecolor{Gray}{gray}{0.9}
\usepackage{mathtools}
\usepackage[T1]{fontenc}
\usepackage[
colorinlistoftodos,
textsize=footnotesize,
]{todonotes} 
 
\usepackage[%
sort&compress
]{cleveref}

\icmltitlerunning{Orthogonal Regularization}

\begin{document}

\twocolumn[
\icmltitle{Estimating Average Treatment Effects via Orthogonal Regularization}

\icmlsetsymbol{equal}{*}

\begin{icmlauthorlist}
	\icmlauthor{Tobias Hatt}{ETH}
	\icmlauthor{Stefan Feuerriegel}{ETH}
\end{icmlauthorlist}

\icmlaffiliation{ETH}{ETH Zurich, Switzerland}

\icmlcorrespondingauthor{Tobias Hatt}{thatt@ethz.ch}

\vskip 0.3in
]

\printAffiliationsAndNotice{} 

\begin{abstract}
	Decision-making often requires accurate estimation of treatment effects from observational data. This is challenging as outcomes of alternative decisions are not observed and have to be estimated. Previous methods estimate outcomes based on unconfoundedness but neglect any constraints that unconfoundedness imposes on the outcomes. In this paper, we propose a novel regularization framework for estimating average treatment effects that exploits unconfoundedness. To this end, we formalize unconfoundedness as an orthogonality constraint, which ensures that the outcomes are orthogonal to the treatment assignment. This orthogonality constraint is then included in the loss function via a regularization. Based on our regularization framework, we develop \textbf{d}eep \textbf{o}rthogonal \textbf{n}etworks for \textbf{u}nconfounded \textbf{t}reatments (DONUT), which learn outcomes that are orthogonal to the treatment assignment. Using a variety of benchmark datasets for estimating average treatment effects, we demonstrate that DONUT outperforms the state-of-the-art substantially.
\end{abstract}

\section{Introduction}
Estimating the causal effect of a treatment is an important problem in many domains such as medicine \citep{hatt2021sequential}, economics \citep{Heckman1997}, and epidemiology \citep{Robins2000a}. For instance, during an epidemic such as the coronavirus disease 2019 (COVID-19), policy makers take unprecedented measures to control the outbreak of the disease \citep[\eg,][]{hellewell2020feasibility, kucharski2020early, prem2020effect}. In order to support evidence-based policy making, it is crucial to estimate the population-wide causal effect of a measure (treatment) on the infection rate (outcome).





The causal effect of a treatment can be estimated in two ways: \emph{randomized control trials (RCTs)} and \emph{observational studies}. RCTs are widely recognized as the gold standard for estimating average causal effects, yet conducting RCTs is often infeasible \citep{Robins2000a}. For instance, randomly allocating different policy measures during an epidemic might be unethical and impractical. Unlike RCTs, observational studies adopt observed data to estimate causal effects. To this end, \emph{unconfoundedness} is required. That is, the covariates $X$ must contain all confounders (\ie, variables that affect both treatment assignment and outcome) such that, when conditioned on $X$, the treatment assignment $T$ is independent of the corresponding outcome $Y(t)$, \ie, $Y(t)\, \indep\, T \mid X$. This is becoming increasingly common due to ease of access to rich data \citep[\eg,][]{Johansson2016, bertsimas2017personalized}. In this paper, we estimate the average causal effect of a treatment from observational data.

In order to estimate the average effect of a treatment from observational data, the outcome of an alternative treatment has to be estimated. However, this is challenging, since we do not know what the outcome would have been if another treatment had been applied. Most existing methods for estimating the average treatment effect can be divided into three categories (see \Cref{sec:realted_work} for an overview). (i)~Regression-based approaches use the treatment assignment as a feature and train regression models to estimate the outcomes \citep[\eg,][]{Funk2011, Wager2018a}, which has been extended to deep learning \citep[\eg,][]{Shalit2017a,Shi2019a}. (ii)~Weighting-based approaches re-weight the data such that the re-weighted distribution resembles the distribution of interest \citep[\eg,][]{Kang2007, austin2011introduction}. (iii)~Doubly-robust approaches seek to combine the first two approaches in a manner that is robust to model misspecification \citep[\eg,][]{Benkeser2017, Chernozhukov2017a}. The aforementioned approaches are widely based on unconfoundedness (\ie, $Y(t)\, \indep\, T \mid X$), which is assumed for identifiability of the treatment effect. Yet, in the aforementioned approaches, during estimation of the model parameters, any implications on the outcomes that arise from unconfoundedness are neglected.

In this paper, we estimate the average treatment effect (ATE). Estimating the ATE is typically based on unconfoundedness. Yet, the constraints that unconfoundedness imposes on the outcomes were never leverages during the estimation of the model parameters. Based on this, we develop a novel regularization framework that exploits unconfoundedness during the estimation of the model parameters. 
We formalize unconfoundedness as an \emph{orthogonality constraint} that requires the outcomes to be orthogonal to the treatment assignment (with respect to some inner product). In order to ensure that the orthogonality constraint holds during the estimation of the model parameters, we accommodate our regularization framework with the orthogonality constraint using a pseudo outcome for the unobserved outcomes and perturbating the expected outcome. While unregularized approaches cannot learn anything about the unobserved outcome, our regularization framework exploits unconfoundedness to learn unobserved outcomes that are orthogonal to the treatment assignment. As such, we leverage additional information about the unobserved outcomes which improves the estimation of the ATE. Based on our regularization framework, we develop \textbf{d}eep \textbf{o}rthogonal \textbf{n}etworks for \textbf{u}nconfounded \textbf{t}reatments (\textbf{DONUT}) for estimating the ATE. DONUT leverages the predictive capabilities of neural networks to learn outcomes that are orthogonal to the treatment assignment. Using a variety of benchmark datasets for estimating treatment effects, we demonstrate that DONUT outperforms the state-of-the-art substantially.

We summarize our \textbf{contributions}\footnote{Code available at \url{github.com/tobhatt/donut}.} as follows:
\begin{enumerate}
	\item We develop a regularization framework for estimating the ATE, which exploits unconfoundedness during the estimation of the model parameters. 
	\item We formalize unconfoundedness as an orthogonality constraint. This constraint ensure that the outcomes are orthogonal to the treatment assignment.
	\item Based on our regularization framework, we use neural networks to develop DONUT and demonstrate that DONUT outperforms the state-of-the-art substantially. 
\end{enumerate}
\section{Related Work}\label{sec:realted_work}
We discuss methods for estimating treatment effects that relate to our proposed method. In particular, we briefly discuss standard statistical methods for estimating the ATE and recent adaptions of machine learning methods. Note that all of these approaches assume some form of unconfoundedness.

A common approach to estimating treatment effects is covariate adjustment \citep{robins1986new, pearl2009causality}, which is also called regression-based approach. Covariate adjustment uses the treatment assignment as well as the covariates as features in order to estimate the expected outcomes \citep[\eg,][]{athey2017estimating, Funk2011, kallus2017recursive}. Extensive works in statistics focus on asymptotic consistency of treatment effect estimates obtained by covariate adjustment \citep{belloni2014inference, Chernozhukov2017a}. Another widely adopted statistical approach to ATE estimation are weighting-based approaches, which seek weights based on the treatment and covariate data to re-weight the subjects in an observational study \citep{athey2016approximate, Kang2007, kallus2018balanced}. For instance, inverse propensity score weighting re-weights the subjects with the inverse of the propensity score so that the re-weighted subjects look like as if they had received an alternative treatment \citep{austin2011introduction}. It is well known that, if the treatment and control group differ substantially, dividing by propensities leads to high variance estimates \citep{swaminathan2015batch}. Hence, additional strategies such as clipping the probabilities away from zero and normalizing by the sum of weights as a control variate are usually required for good performance \citep{swaminathan2015self, wang2017optimal}. Doubly-robust approaches go further and combine covariate adjustment with propensity score weighting in a manner that is robust to model misspecification \citep{athey2017estimating, van2006targeted, Funk2011, Benkeser2017, Chernozhukov2017a}.

In recent years, many machine learning methods have been adapted for estimating treatment effects \citep{cui2020causal, hatt2021sequential}. For instance, random forest-based methods estimate the treatment effect at the leaf node and train many weak learners to build expressive ensemble models \citep{Wager2018a}. Deep learning has been successful for the task of estimating treatment effects due to its strong predictive performance and ability to learn representations of the data \citep[\eg,][]{Johansson2016,Louizos2017, rakesh2018linked, Shalit2017a, Yao2018a, Yoon2018a, Shi2019a, chu2020matching, liu2020estimating, berrevoets2020organite}. For instance, \citep{Yoon2018a} use generative adversarial networks to generate the outcomes of alternative treatments such that they cannot be distinguished from the observed outcomes by the discriminator. \citep{Johansson2016} proposed learning a representation of the data such that the treated and control group are more similar. \citep{Shalit2017a} built upon this idea, but base their algorithm on a neural network architecture, which they termed Treatment-Agnostic Representation Network (TARNet). \citep{Shi2019a} extend the architecture of TARNet to estimate the average treatment effect (termed Dragonnet). In particular, Dragonnet provides an end-to-end procedure for simultaneously estimating the propensity score and the expected outcomes from the covariates and treatment assignment. The simultaneous estimation leads to a trade off between prediction quality of the outcomes and a good representation of the propensity score, which improves the estimation of the ATE. Bayesian algorithms have been adapted for estimating the ATE due to their ability to quantify uncertainty. Specifically, \citep{Ray2019a} propose a data-driven modification to the prior based on the propensity score that corrects for the first-order posterior bias for the ATE, which improves performance for both estimation accuracy and uncertainty quantification.

All of the above methods are based on unconfoundedness, which is assumed for identifiability of the treatment effect from observational data. However, during estimation of the model parameters, any implications on the outcomes that arise from unconfoundedness have been neglected. We address this shortcoming by formalizing unconfoundedness as an orthogonality constraint, which is then included in the estimation of the model parameters.



\section{Problem Setup}
Our objective is to estimate the average treatment effect (ATE) of a binary treatment from observational data. For this, we build upon the Neyman-Rubin potential outcomes framework \citep{Rubin2005}. Consider a population where every subject $i$ is described by the $d$-dimensional covariates $X_i \in \mathcal{R}^d$. Each subject is assigned a treatment $T_i \in \{0,1\}$. The random variable $Y_i(1)$ corresponds to the outcome under treatment, \ie, $T_i=1$, whereas $Y_i(0)$ corresponds to the outcome under \emph{no} treatment, \ie, $T_i=0$. These two random variables, $Y_i(1), Y_i(0) \in \Rl$, are known as the potential outcomes. Due to the fundamental problem of causal inference, only one of the potential outcomes is observed, but never both. The observed outcome is denoted by $Y_i$. 

Our aim is to estimate the \emph{average treatment effect}
\begin{equation}\label{ATE}
	\psi = \Efb{Y(1) - Y(0)}.
\end{equation}
The following standard assumptions are sufficient for identifiability of the treatment effect from observational data \citep{imbens2015causal}.

\textbf{Assumption 1}~(\emph{Consistency}.) \textit{The potential outcome of treatment $T$ equals the observed outcome if the actual treatment received is $T$, \ie, $\forall t\in\{0,1\}: Y=Y(t)$, if $T=t$.} 

\textbf{Assumption 2}~(\emph{Positivity}.) \textit{For any set of covariates $X$, the probability to receive treatment $0$ or $1$ is positive, \ie, $\forall x \in \mathcal{R}^d: 0 < \Prb{T=1 \mid X=x} < 1$.}

\textbf{Assumption 3}~(\emph{Unconfoundedness}.) \textit{The potential outcomes $Y(1)$ and $Y(0)$ are independent of the treatment assignment $T$ given the covariates $X$, \ie,}
\begin{equation}
	Y(1), Y(0)\, \indep\, T \mid X.
\end{equation}
Assumption~3 is also called ``no unmeasured confounders'', or ``(strong) ignorability''.

Based on this, the ATE is equal to
\begin{equation}
	\psi = \Efb{\Efb{Y \mid X,\,T=1} - \Efb{Y \mid X,\,T=0}}.
\end{equation}
Then, our aim is to estimate the function $f(x, t) = \Efb{Y \mid X=x,\,T=t}$ for all $x\in\mathcal{R}^d$ and $t\in\{0,1\}$ based on observational data $\dat = \{(X_i, T_i, Y_i)\}_{i=1}^n$.

\section{Orthogonal Regularization for Estimating Average Treatment Effects}
The key idea of our regularization framework is to exploit the implications on the outcomes that result from unconfoundedness. For this, we develop a regularizer which accommodates unconfoundedness as a orthogonality constraint. This orthogonality constraint ensures that the outcomes are orthogonal to the treatment assignment, which is required by unconfoundedness. We later introduce a specific variant of our regularization framework based on neural networks, which we name DONUT.

\subsection{Unconfoundedness as Orthogonality Constraint}
Under unconfoundedness, the outcomes are independent of the treatment assignment given the covariates, \ie,
\begin{equation}\label{eq:NUC}
	Y(1), Y(0)\, \indep\, T \mid X.
\end{equation}

Using the inner product $\langle V, W \rangle = \frac1n \sm i n{V_i\, W_i}$, we formalize the following \emph{orthogonality constraint}:
\begin{align}\label{eq:orthogonality}
	&\langle Y(t) - f(X, t), T - \pi(X)\rangle \\
	&= \frac1n \sm i n (Y_i(t) - f(X_i, t))(T_i - \pi(X_i)) = 0,
\end{align}
for $t\in\{0,1\}$, and where $f(x, t) = \Efb{Y \mid X=x,\, T=t}$ and $\pi(x) = \Efb{T\mid X=x}$. The function $\pi$ is the \emph{propensity score}.\footnote{Note that the propensity score is defined as $\pi(x) = \Prb{T=1\mid X=x}$ \citep{Trust2016}, which is equivalent to $\Efb{T\mid X=x}$ for binary treatments.} The orthogonality constraint is a necessary condition for unconfoundedness. Unconfoundedness requires the conditional covariance between $Y(t)$ and $T$ to be zero, since they are conditionally independent. Therefore, the inner product in \labelcref{eq:orthogonality} also has to be zero, since it is the empirical expected conditional covariance between $Y(t)$ and $T$. As a result, the orthogonality constraint requires the (centered) outcomes to be orthogonal to the (centered) treatment assignment with respect to the above inner product.\footnote{Note that we use the notion of orthogonality with respect to the inner product in \labelcref{eq:orthogonality}. This is different from the notion of Neyman orthogonality in \citep{nie2017quasi, Chernozhukov2017a}, which requires the G$\hat{\text{a}}$teau derivative of a debiased score function to vanish at the true parameter. In contrast, our orthogonality constraint exploits unconfoundedness to ensure that the outcomes are orthogonal (w.r.t. the inner product in \labelcref{eq:orthogonality}) to the treatment assignment.} An appropriate method for estimating outcomes based on unconfoundedness should ensure that the orthogonality constraint holds. Although existing methods are based on unconfoundedness, the orthogonality of the outcomes to the treatment assignment is ignored during estimation of the model parameters. As a remedy, we propose a regularization framework that accommodates the orthogonality constraint in the estimation procedure.
\subsection{Proposed Regularization Framework}\label{sec::proposed_framework}
In our regularization framework, the orthogonality constraint in \labelcref{eq:orthogonality} is included in the estimation procedure as follows. Let $\mathcal H_f\subseteq [\mathcal R^d \times \fbraces{0,1} \rightarrow \Rl]$ and $\mathcal H_\pi\subseteq [\mathcal R^d \rightarrow \fbracks{0,1}]$ be function classes for $f$ and $\pi$, and $\epsilon \in \Rl$ a model parameter. Then, the objective is to find the solution to the optimization problem
\begin{equation}\label{eq:OptProblem}
	\inf_{\substack{(f, \pi, \epsilon) \in \mathcal H_f\times \mathcal H_\pi\times \Rl}} \mathcal{L}_{\mathit{FL}}(f, \pi; \dat) + \lambda\,\Omega_{\mathit{OR}}(f,\pi,\epsilon; \dat),
\end{equation}
where $\mathcal{L}_{\mathit{FL}}$ is the so-called factual loss (see \Cref{sec:Factual_Loss}) between the estimated models and observed (factual) data. The term $\Omega_{\mathit{OR}}$ is our new orthogonal regularization that enforces the orthogonality constraint using the model parameter $\epsilon$ (see \Cref{sec::counterfactual_regularization}). The variable $\lambda\in\Rl_{+}$ is a hyperparameter controlling the strength of regularization. We describe both components of \labelcref{eq:OptProblem}, \ie, factual loss and orthogonal regularization, in the following.

\subsubsection{Factual Loss $\mathcal{L}_{\mathit{FL}}$}\label{sec:Factual_Loss}
Similar to previous work \citep[\eg,][]{Shi2019a}, the outcome model $f$ and propensity score model $\pi$ are estimated using the observed data and the factual loss given by
\begin{equation}
	\begin{aligned}\label{eq:FactualLoss}
		\mathcal{L}_{\mathit{FL}}(f, \pi; \dat) = \frac{1}{n}&\sm i n (f(X_i, T_i) - Y_i)^2 \\&+ \alpha\, \text{CrossEntropy}(\pi(X_i), T_i),
	\end{aligned}
\end{equation}
where $\alpha\in\Rl_{+}$ is a hyperparameter weighting the terms of the factual loss. We note the following about the estimation of the outcome model $f$ in \labelcref{eq:FactualLoss}. In the first term of the factual loss, $f$ is fitted to the observed outcomes and, thus, no information about the unobserved outcomes is used. As a consequence, the model learns about the observed outcomes, but not about the unobserved outcomes. State-of-the-art methods \emph{exclusively} rely on the factual loss to estimate the outcomes \citep[\eg][]{Shalit2017a}. This is based on the assumption that similar subjects have similar outcomes \citep{Yao2018a} and that, for every subject in the data, there is a similar subject in the data with the opposite treatment. However, this is unlikely in presence of selection bias, \ie, when treatment and control groups differ substantially. This pitfall is addressed by our orthogonal regularization in the following.

\subsubsection{Orthogonal Regularization $\Omega_{\mathit{OR}}$}\label{sec::counterfactual_regularization}
We now specify the orthogonal regularization term $\Omega_{\mathit{OR}}$ that ensures that the orthogonality constraint in \labelcref{eq:orthogonality} is satisfied. We first state $\Omega_{\mathit{OR}}$ and then explain each of its parts. The orthogonal regularization term is given by
\begin{equation}\label{eq:regularizer}
	\Omega_{\mathit{OR}}(f, \pi, \epsilon; \dat) = \frac{1}{n}\sm i n (Y_i^\ast(0) - f^\epsilon(X_i, 0))^2,
\end{equation}
with
\begin{equation}\label{eq:ATEproxy}
	Y_i^\ast(0) = Y_i - \psi^\ast\,T_i,\quad \psi^\ast = \frac1n \sm i n{f(X_i, 1) - f(X_i, 0)}
\end{equation}
and 
\begin{equation}\label{eq:orthogonality_enforce}
	f^\epsilon(X_i, t) = f(X_i, t) + \epsilon(T_i - \pi(X_i)),
\end{equation}
for $t\in\{0,1\}$, where the pseudo outcome $Y_i^\ast(0)$ and the perturbation function $f^\epsilon$ are required to learn outcomes that are orthogonal to the treatment assignment, as described in the following.

\textbf{Pseudo outcome.} The untreated outcome of a subject can be expressed as $Y_i(0) = Y_i - \psi_i\,T_i$, where $\psi_i = Y_i(1) - Y_i(0)$ is the true treatment effect of subject $i$. Hence, if we had access to the treatment effect $\psi_i$, we would also have access to the untreated outcome $Y_i(0)$ even if we did not observe it.\footnote{This can be seen by distinction of cases: if $T_i=1$, then $Y_i=Y_i(1)$ and, therefore, $Y_i(0) = Y_i - \psi_i$, and if $T_i=0$, then $Y_i(0) = Y_i$.} We use the average treatment effect of the current model fit, \ie, $\psi^\ast$ in \labelcref{eq:ATEproxy}, as proxy for $\psi_i$. This creates the pseudo outcome, $Y_i^\ast(0) = Y_i - \psi^\ast\,T_i$.

\textbf{Perturbation function.} In order to ensure that the orthogonality constraint is satisfied, we use a perturbation function $f^\epsilon$. Simply adding the inner product as a regularization term is not sufficient, since this does not guarantee that the orthogonality constraint holds at a solution of \labelcref{eq:OptProblem}. A similar approach is used in targeted minimum loss estimation \citep[\eg,][]{van2011targeted} and, more general, dual problems in optimization \citep{zalinescu2002convex}. We extend the function $f$ to a perturbation function $f^\epsilon$ using the model parameter $\epsilon$ as in \labelcref{eq:orthogonality_enforce}. As a result, solving the optimization problem in \labelcref{eq:OptProblem} forces the outcome estimates $f^\epsilon$ to satisfy the orthogonality constraint. Mathematically, this can be seen by taking the partial derivative of \labelcref{eq:OptProblem} and setting it to zero, \ie,
\begin{align}\label{eq:derivative_epsilon}
	0 &= \frac{\partial}{\partial \epsilon} (\mathcal{L}_{\mathit{FL}} + \lambda\,\Omega_{\mathit{OR}}(f, \pi, \epsilon; \dat))\Big\vert_{\epsilon = \hat{\epsilon}}\\
	&= 2\frac{1}{n}\sm i n (Y_i^\ast(0) - f^{\hat{\epsilon}}(X_i, 0))\,(T_i - \pi(X_i)),
\end{align} 
and, hence, $\langle Y^\ast(0) - f^\epsilon(X, 0), T - \pi(X)\rangle=0$. It is analytically sufficient to regularize only the untreated outcome.\footnote{Whenever one of the outcomes is orthogonal to the treatment assignment, the other outcome is also orthogonal. However, it is straightforward to extend the orthogonal regularization term to incorporate both outcomes by adding the same term for the treated outcome $Y(1)$, including the pseudo outcome $Y^\ast(1)$ and the perturbation function $f^\epsilon(x, 1)$ to enforce orthogonality. See \Cref{apx:OnlyTreatmentfree} for details. Incorporating both outcomes into the regularization framework did not yield an improvement in ATE estimation in the experiments.} 

Then, the outcomes are estimated using $f^\epsilon(x, t)$ and, therefore, the estimate for the ATE is $\hat\psi = \frac1n \sm i n \hat{f^\epsilon}(X_i, 1) - \hat{f^\epsilon}(X_i, 0)$, which coincides with $\psi^\ast$, since the perturbation terms in \labelcref{eq:orthogonality_enforce} cancel.

\subsection{Intuition: Why Orthogonal Regularization improves estimation of ATE}
We give an intuition on why exploiting unconfoundedness in the estimation of the model parameters improves estimation of the ATE. State-of-the-art methods are typically based on the factual loss in \labelcref{eq:FactualLoss}. This loss only fits the factual (observed) outcome and, hence, existing methods cannot learn anything about the unobserved outcomes. This can be particularly harmful if the treatment and control group differ substantially, since, in this case, for a treated subject, we cannot find a similar untreated subject and vice versa. Hence, since no information about the untreated outcome is available, existing methods yield poor estimates of the untreated outcome and, therefore, poor estimates of the ATE. Contrary to that, our regularization framework incorporates information about the unobserved outcomes by restricting the unobserved outcomes to be orthogonal to the treatment assignment. As a result, even if the treatment and control group differ substantially, our framework can learn about the unobserved outcome, which provides better estimation of the unobserved outcomes and, hence, yields improved estimation of the ATE.

\subsection{Deep Orthogonal Networks for Unconfounded Treatments}
Our regularization framework works with any outcome model $f$ and propensity score model $\pi$ that is estimated via a loss function. The regularization framework ensures that the outcomes are estimated such that they are orthogonal to the treatment assignment. We introduce a specific variant of our regularization framework based on neural networks, which yields \textbf{d}eep \textbf{o}rthogonal \textbf{n}etworks for \textbf{u}nconfounded \textbf{t}reatments (\textbf{DONUT}). Neural networks present a suitable model class due to their strong predictive performance. The architecture of DONUT is set as follows. For the outcome model $f$, we use the basic architecture of TARNet \citep{Shalit2017a}. TARNet uses a feedforward neural network to produce a representation layer, followed by two 2-layer neural networks to predict the potential outcomes from the shared representation. For the propensity score model $\pi$, we use a logistic regression \citep{Trust2016}.

\subsection{Theoretical Results}\label{sec:theory}
In this section, we give sufficient conditions under which our regularization framework yields an estimator $\hat\psi$ that converges to the true ATE $\psi$ at a rate $1/\sqrt{n}$. Since the conditions may be restrictive in practice, we emphasize that our method does \emph{not} rely on these theoretical results, but we state the result for completeness of our work. We empirically confirm the capability of our method in estimating the ATE in \Cref{sec:experiments}. 
\begin{proposition}\label{thm:asymptotic_linearity}(Convergence.)
	Suppose
	\vspace{-0.5em}
	\begin{enumerate}
		\item The estimator $\hat{\eta} = (\hat{f}, \hat{\pi})$ for the outcome and propensity score model in \labelcref{eq:OptProblem} converges to some $\bar{\eta} = (\bar{f}, \bar{\pi})$ in the sense that $\lVert \hat{\eta} - \bar{\eta} \rVert = o_p(1/\sqrt{n})$, where either $\bar{f} = f$ or $\bar{\pi} = \pi$ (or both) corresponds to the true function.
		\item The treatment effect is homogeneous, \ie, $\forall x\in \mathcal{R}^d: \psi(x)=\psi$.
		\item The estimators $\hat{f}$ and $\hat{\pi}$ take values in $\mathbb P$-Donsker classes, \ie, $\mathcal H_f, \mathcal H_\pi \in \text{CLT}(\mathbb{P})$.\footnote{A class $\mathcal F$ of measurable functions on a probability space $\fprns{\Omega, \mathcal A, \mathbb P}$ is called a $\mathbb{P}$-Donsker class if, for $\mathbb G_n^{\mathbb P} = \sqrt{n}\fprns{\mathbb P_n - \mathbb P}$, the empirical process $\{\mathbb G_n^{\mathbb P}f: f\in\mathcal F\}_{n\geq 1}$ converges weakly to a $\mathbb{P}$-Brownian bridge $\mathbb G^{\mathbb P}$.
			Donsker classes include parametric classes, but also many other classes, including infinite-dimensional classes, \eg, smooth functions and bounded monotone functions. See \citep{Mikosch1997} for more details.}
	\end{enumerate}
	Then, $\hat\psi$ is asymptotically normal, \ie, 
	\vspace{-0.25em}\begin{equation}
		\sqrt{n}\,(\hat\psi - \psi) \xrightarrow[]{d} \mathcal{N}\Big(0, \frac{\sigma^2}{\Efb{\normalfont\text{Var}(T\mid X)}}\Big),
	\end{equation}
	\vspace{-0.25em}where $\sigma^2$ is the variance of the outcome.\,\, \qed
\end{proposition}
The proof is provided in \Cref{apx:proof:AsymptoticNormality}. Since \Cref{thm:asymptotic_linearity} is not necessary for our method to work, but rather gives conditions under which it is guaranteed to converge, we only briefly discuss the result. We make the following remarks about the result and its conditions. Condition~1 requires that either the model for the outcomes $f$ or the model for the propensity score $\pi$ (or both) is consistent in \labelcref{eq:OptProblem}, but not necessarily both and converge at a fast enough rate. In particular, neural networks converge at a fast enough rate to invoke this condition \citep{Farrell2018}. Condition~2 requires that the treatment effect is homogeneous, which has been seen to be beneficial to reduce the asymptotic variance of an estimator \citep[\eg,][]{Vansteelandt2014a}. This may be too restrictive to hold true in practice, but it is not necessary for our method to work (see \Cref{sec:experiments} for the empirical results). Condition~3 captures a large class of functions of estimators for $f$ and $\pi$ from which we can choose. In particular, the class of feedforward neural networks is a $\mathbb P$-Donsker class, since it is a composition of Lipschitz continuous functions. Together with the convergence rate of neural networks \citep{Farrell2018}, this provides justification for the use of neural networks in our regularization framework, and therefore in DONUT.

\textit{Remark.} (Similarity to partially linear regression.) We find an interesting similarity between our estimator $\hat{\psi}$ and the estimator obtained by partially linear regression (PLR) (see (1.5) in \citep{Chernozhukov2017a}). The analytical expression of $\hat{\psi}$ yielded by our regularization framework in \labelcref{eq:OptProblem} is given in \labelcref{apx:eq:anal_expression} in \Cref{apx:proof:AsymptoticNormality}. This is similar to the estimator obtained by PLR after partialling the effect of $X$ out from $T$ (see (1.5) in \citep{Chernozhukov2017a}). However, the PLR separately estimates the nuisance functions $\hat{f}$ and $\hat{\pi}$ and then plugs them into the estimator. Our approach does not use the analytical expression as plug-in estimator, but the estimator arises directly from solving the optimization problem in \labelcref{eq:OptProblem} via \labelcref{eq:regularizer}. The difference becomes apparent in the experiments (\Cref{sec:experiments}), where we include the PLR as a baseline and find that the performance of DONUT is superior.

\section{Experiments}\label{sec:experiments}
Our proposed method for estimating the ATE, DONUT, is evaluated against state-of-the-art baselines, where we find that its ATE estimation is superior (\Cref{sec:results}). We show in an ablation study the benefit of the orthogonal regularization $\Omega_{OR}$, where we find that it substantially improves the estimation performance (\Cref{sec:ablation_study}). We further demonstrate that DONUT achieves robust results for problems subject to selection bias, which is confirmed as part of a simulation study (\Cref{sec:Simulation_Study}). 

\subsection{Setup}
Evaluating methods for estimating treatment effects is challenging, since we rarely have access to ground truth treatment effects. Established procedures for evaluation of such methods rely on semi-synthetic data, which reflect the real world. Our experimental setup follows established procedure regarding datasets, baselines, and performance metrics \citep[\eg,][]{Johansson2016, Shalit2017a}.

\textbf{Datasets.} We evaluate all methods across three benchmark datasets for estimating treatment effects: \textbf{IHDP} \citep[\eg,][]{Johansson2016}, \textbf{Twins} \citep[\eg,][]{Yoon2018a}, and \textbf{Jobs} \citep[\eg,][]{Shalit2017a}. The first two datasets are semi-synthetic, while the last originated from a randomized experiment. Details on IHDP, Twins, and Jobs are provided in corresponding papers. 

\textbf{Training details.} DONUT is trained using the regularization framework in \labelcref{eq:OptProblem}, where both the outcome model and the propensity score model are trained jointly using stochastic gradient descent with momentum. The hidden layer size is 200 for the representation layers and 100 for the outcome layers similar to \citep{Shalit2017a, Shi2019a}. The hyperparameter $\alpha$ in the factual loss \labelcref{eq:FactualLoss} is set to 1 and $\lambda$ in the orthogonal regularization \labelcref{eq:regularizer} is determined by hyperparameter optimization over $\{10^k\}_{k=-2}^{2}$. For IHDP, we follow established practice \citep[\eg,][]{Shalit2017a} and average over 1,000 realizations of the outcomes with 63/27/10 train/validation/test splits. Following \citep{Shalit2017a, Yoon2018a}, we average over 100 train/validation/test splits for Twins and Jobs, all with ratios 56/24/20.

\textbf{Baselines.} We compare DONUT against 18 state-of-the-art methods for estimating treatment effects, organized in the following groups: (i)~Regression methods: Linear regression with treatment as covariate (\textbf{OLS/LR-1}), separate linear regressors for each treatment (\textbf{OLS/LR-2}), and balancing linear regression (\textbf{BLR}) \citep{Johansson2016}; (ii)~Matching methods: $k$-nearest neighbor (\textbf{$k$-NN}) \citep{Crump2008a}; (iii)~Tree methods: Bayesian additive regression trees (\textbf{BART}) \citep{Chipman2012a}, random forest (\textbf{R-Forest}) \citep{Breiman2001}, and causal forest (\textbf{C-Forest}) \citep{Wager2018a}; (iv)~Gaussian process methods for estimating ATE: debiased Gaussian process (\textbf{D-GP}) \citep{Ray2019a}; (v)~Neural network methods: Balancing neural network (\textbf{BNN}) \citep{Johansson2016}, treatment-agnostic representation network (\textbf{TARNet}) \citep{Shalit2017a}, counterfactual regression with Wasserstein distance (\textbf{CFR-WASS}) \citep{Shalit2017a}, generative adversarial networks (\textbf{GANITE}) \citep{Yoon2018a}, and \textbf{Dragonnet} \citep{Shi2019a}; (vi)~Semiparametric methods: Double robust estimator (\textbf{DRE}) \citep{athey2017estimating}, targeted maximum likelihood estimator (\textbf{TMLE}) \citep{van2006targeted}, and partially linear regression (\textbf{PLR}) \citep{Chernozhukov2017a}, all of them using the outcome and propensity score model of Dragonnet as nuisance functions, and Double selection estimator (\textbf{DSE}) \citep{athey2017estimating} and approximate residual re-balancing estimator (\textbf{ARBE}) \citep{athey2016approximate}.

\textbf{Performance metrics.} Following established procedure, we report the following metrics for each dataset. For IHDP, we use the absolute error in average treatment effect \citep{Johansson2016}: $\epsilon_\text{ATE} = \vert\frac{1}{n}\sm i n(f(x_i, 1) - f(x_i, 0)) - \frac{1}{n}\sm i n(\hat{f}(x_i, 1) - \hat{f}(x_i, 0))\vert$. For Twins, we use the absolute error in observed average treatment effect \citep{Yoon2018a}: $\epsilon_\text{ATE} = \vert\frac{1}{n}\sm i n(y_i(1) - y_i(0)) - \frac{1}{n}\sm i n(\hat{y}_i(1) - \hat{y}_i(0))\vert$. For Jobs, all treated subjects $\mathcal{T}$ were part of the original randomized sample $\mathcal{E}$, and hence, the true average treatment effect can be computed on the treated by $\text{ATT} = \vert \mathcal{T}\vert^{-1}\sum_{i \in \mathcal{T}} y_i - \vert\mathcal{C} \cap \mathcal{E}\vert^{-1}\sum_{i \in \mathcal{C} \cap \mathcal{E}} y_i$, where $\mathcal{C}$ is the control group. Similar to \citep{Shalit2017a}, we then use the absolute error in average treatment effect on the treated: $\epsilon_\text{ATT} = \vert \text{ATT} - \vert\mathcal{T}\vert^{-1}\sum_{i \in \mathcal{T}} (\hat{f}(x_i, 1) - \hat{f}(x_i, 0))\vert$.
\subsection{Results}\label{sec:results}
\begin{table*}[t]
	\caption{\label{tbl:ihdp_twins_jobs_results}Results for estimating average treatment effects on IHDP, Twins, and Jobs. Lower is better.}
	\begin{center}
		\begin{small}
			\begin{sc}
				\begin{tabular}{l@{\hspace*{10mm}}cc@{\hspace*{7.5mm}}cc@{\hspace*{7.5mm}}cc}
					\multicolumn{4}{l}{\bf{Results}} \\
					\toprule
					&\multicolumn{6}{c}{\bf{Datasets (Mean $\pm$ Std)}}\\
					\cmidrule{2-7}
					&\multicolumn{2}{c}{\textbf{IHDP} ($\epsilon_\text{ATE}$)}& \multicolumn{2}{c}{\textbf{Twins} ($\epsilon_\text{ATE}$)} & \multicolumn{2}{c}{\textbf{Jobs} ($\epsilon_\text{ATT}$)} \\
					Method & In-s. & Out-s. & In-s. & Out-s. & In-s. & Out-s.\\
					\hline
					\addlinespace[0.75ex]
					\textsc{OLS/LR-1} & $.73 \pm .04$ & $.94 \pm .06$ & $.0038 \pm .0025$ & $.0069 \pm .0056$ & $.01 \pm .00$ & $.08 \pm .04$ \\
					\textsc{OLS/LR-2} & $.14 \pm .01$ & $.31 \pm .02$ & $.0039 \pm .0025$ & $.0070 \pm .0059$ & $.01 \pm .01$ & $.08 \pm .03$ \\
					\textsc{BLR} & $.72 \pm .04$ & $.93 \pm .05$ & $.0057 \pm .0036$ & $.0334 \pm .0092$ & $.01 \pm .01$ & $.08 \pm .03$ \\
					\textsc{$k$-NN} & $.14 \pm .01$ & $.79 \pm .05$ & $.0028 \pm .0021$ & $.0051 \pm .0039$ & $.21 \pm .01$ & $.13 \pm .05$ \\
					\textsc{BART} & $.23 \pm .01$ & $.34 \pm .02$ & $.1206 \pm .0236$ & $.1265 \pm .0234$ & $.02 \pm .00$ & $.08 \pm .03$ \\
					\textsc{R-Forest} & $.73 \pm .05$ & $.96 \pm .06$ & $.0049 \pm .0034$ & $.0080 \pm .0051$ & $.03 \pm .01$ & $.09 \pm .04$ \\
					\textsc{C-Forest} & $.18 \pm .01$ & $.40 \pm .03$ & $.0286 \pm .0035$ & $.0335 \pm .0083$ & $.03 \pm .01$ & $.07 \pm .03$ \\
					\textsc{BNN}& $.37 \pm .03$ & $.42 \pm .03$ & $.0056 \pm .0032$ & $.0203 \pm .0071$ & $.04 \pm .01$ & $.09 \pm .04$ \\
					\textsc{TARNet} & $.26 \pm .01$ & $.28 \pm .01$ & $.0108 \pm .0017$ & $.0151 \pm .0018$ & $.05 \pm .02$ & $.11 \pm .04$ \\
					\textsc{CFR-WASS}& $.25 \pm .01$ & $.27 \pm .01$ & $.0112 \pm .0016$ & $.0284 \pm .0032$ & $.04 \pm .01$ & $.09 \pm .03$ \\
					\textsc{GANITE} & $.43 \pm .05$ & $.49 \pm .05$ & $.0058 \pm .0017$ & $.0089 \pm .0075$ & $.01 \pm .01$ & $.06 \pm .03$ \\
					\textsc{DRE} & $.13 \pm .12$ & $.22 \pm .28$ & $.0027 \pm .0013$ & $.0048 \pm .0032$ &$ .03 \pm .01$ & $.10 \pm .08$\\
					\textsc{DSE} & $.21 \pm .17$ & $.27 \pm .35$ &$.0261 \pm .0165$  & $.0227 \pm .0157$ & $.06 \pm .05$& $.18 \pm .13$\\
					\textsc{ARBE} & $.21 \pm .10$ & $.61 \pm .39$ &$.0225 \pm .0115$  & $.0223 \pm .0147$ & $.06 \pm .05$& $.17 \pm .12$\\
					\textsc{PLR} & $.63 \pm .09$ & $1.32 \pm .31$ & $.0133 \pm .0255$ & $.0084 \pm .0354$ &$ .09 \pm .04$ & $.10 \pm .08$\\
					\textsc{TMLE} & $.13 \pm .10$ & $.33 \pm .32$ & $.0034 \pm .0020$ & $.0053 \pm .0025$ &$ .02 \pm .01$ & $.06 \pm .05$\\
					\textsc{D-GP} & $.14 \pm .32$ & $.17 \pm .47$ &$.0046 \pm .0033$ &$.0058 \pm .0043$ & $.03 \pm .01$ & $.06 \pm .05$ \\
					\textsc{Dragonnet} & $.14 \pm .01$ & $.20 \pm .05$ & $.0062 \pm .0051$ & $.0064 \pm .0054$ &$ .02 \pm .01$ & $.06 \pm .05$\\
					\hline
					\addlinespace[0.75ex]
					\textbf{\textsc{DONUT}} & $\bf{.13} \pm \bf{.01}$ & $\bf{.19} \pm \bf{.02}$ &$\bf{.0025} \pm \bf{.0016}$ & $\bf{.0033} \pm \bf{.0026}$ & $\bf{.01} \pm \bf{.00}$ & $\bf{.06} \pm \bf{.05}$ \\					
					\bottomrule	
					\\
				\end{tabular}
			\end{sc}
		\end{small}
	\end{center}
	\vspace{-1em}
\end{table*}
IHDP, Twins, and Jobs have been used to evaluate many methods for estimating treatment effects. In \Cref{tbl:ihdp_twins_jobs_results}, the results of the experiments on IHDP, Twins, and Jobs are presented\footnote{The numbers for the baselines are taken from the corresponding papers if possible.}. Overall, DONUT achieves superior performance across all datasets. 

On IHDP, the D-GP baseline achieves slightly worse in-sample estimation error (mean: .14 (D-GP) vs. .13 (DONUT)) and slightly better out-of-sample estimation error (mean: .17 (D-GP) vs. .19 (DONUT)), but at the drawback of an inferior standard deviation (std.: .47 (D-GP) vs. .02 (DONUT)). However, as shown by previous work, the small sample size and the limited simulation settings of IHDP make it difficult to draw conclusions about methods \citep[\eg,][]{Yoon2018a, Shi2019a}. On Twins and Jobs, where the number of samples is much larger (Twins: $n=$11,400; Jobs: $n=$3,212), DONUT achieves superior performance. On these datasets, DONUT is state-of-the-art and outperforms all baselines for estimating ATE from the statistical literature (\eg, TMLE or ARBE) and all state-of-the-art baselines for estimating ATE from the machine learning literature (D-GP and Dragonnet). Similar to \citep{Shi2019a}, we also evaluated DONUT on the ACIC 2018 dataset collection, where we find that, across a large collection of datasets and methods, DONUT achieves superior results. The results can be found in \Cref{sec:acic_2018}. The hyperparameter $\lambda$ controlling the strength of regularization is determined by hyperparameter optimization. We conduct a sensitivity analysis and find that the results are robust. The corresponding analysis can be found in \Cref{apx:sensitivity_lambda}.

Among neural network-based methods (\ie, BNN, TARNet, CFR-WASS, GANITE, Dragonnet, and DONUT), DONUT performs superior across all datasets. In particular, the current state-of-the-art methods, Dragonnet, is outperformed by DONUT across all datasets. In comparison to TARNet, which shares the basic architecture of DONUT, we achieve substantial reduction in ATE estimation error across all datasets (out-of-sample error reduction: 32.1\,\% on IHDP, 78.1\,\% on Twins, and 45.5\,\% on Jobs). This demonstrates the effectiveness of our regularization framework. 
\begin{table*}[t]
	\caption{\label{tbl:ablation_study}Results of ablation study for estimating average treatment effects. Comparison between DONUT and DONUT w/o $\Omega_{OR}$ (which is DONUT, but where the orthogonal regularizer is removed). All other experimental variables are fixed. Lower is better.}
	\begin{center}
		\begin{small}
			\begin{sc}
				\begin{tabular}{lcccccc}
					\multicolumn{4}{l}{\bf{Ablation study}} \\
					\toprule
					&\multicolumn{6}{c}{\bf{Datasets (Mean $\pm$ Std)}}\\
					\cmidrule{2-7}
					&\multicolumn{2}{c}{\textbf{IHDP} ($\epsilon_\text{ATE}$)}& \multicolumn{2}{c}{\textbf{Twins} ($\epsilon_\text{ATE}$)} & \multicolumn{2}{c}{\textbf{Jobs} ($\epsilon_\text{ATT}$)} \\
					\addlinespace[0.75ex]
					Method & In-s. & Out-s. & In-s. & Out-s. & In-s. & Out-s.\\
					\hline
					\addlinespace[0.75ex]
					\textsc{DONUT w/o $\Omega_{OR}$} & $.23 \pm .01$ & $.31 \pm .02$ & $.0069 \pm .0053$ & $.0080 \pm .0058$ &$.02 \pm .01$ & $.10 \pm .05$\\
					\hline
					\addlinespace[0.75ex]
					\textbf{\textsc{DONUT}} & $\bf{.13} \pm \bf{.01}$ & $\bf{.19} \pm \bf{.02}$ &$\bf{.0025} \pm \bf{.0016}$ & $\bf{.0033} \pm \bf{.0026}$ & $\bf{.01} \pm \bf{.00}$ & $\bf{.06} \pm \bf{.05}$ \\					
					\bottomrule	
					\\
				\end{tabular}
			\end{sc}
		\end{small}
	\end{center}
\end{table*}
\subsection{Ablation Study on Orthogonal Regularizer $\Omega_{OR}$}\label{sec:ablation_study}
We further evaluate the contribution of the orthogonal regularizer $\Omega_{OR}$ to the performance of DONUT in an ablation study. To this end, we compare DONUT with the orthogonal regularizer $\Omega_{OR}$ (denoted as DONUT) against DONUT without the orthogonal regularizer $\Omega_{OR}$ (denoted as DONUT w/o $\Omega_{OR}$). All other experimental variables are fixed. \Cref{tbl:ablation_study} presents the results of the experiments on IDHP, Twins, and Jobs. The main observation is that DONUT consistently improves estimation relative to DONUT w/o $\Omega_{OR}$ (DONUT without orthogonal regularizer) across all datasets. Specifically, the orthogonal regularizer $\Omega_{OR}$ improves the out-of-sample estimation error by $38.7\%$ on IHDP, $62.5\%$ on Twins, and $39.7\%$ on Jobs. This result further confirms the benefit of including orthogonal regularization in the estimation procedure.

\subsection{Simulation Study on Selection Bias}\label{sec:Simulation_Study}
We evaluate the robustness of DONUT with regards to selection bias (\ie, when treatment and control groups differ substantially), using simulated data with varying selection bias generated according to a similar protocol as in \citep{Yao2018a, Yoon2018a}. We generate 2,500 untreated samples from $\mathcal N(\mathbf{0}^{10\times1}, 0.5\times\Sigma\Sigma^\top)$ and 5,000 treated samples from $\mathcal N(\mu_1, 0.5\times\Sigma\Sigma^\top)$, where $\Sigma\sim\mathcal U((0,1)^{10\times10})$. Varying $\mu_1$ yields different levels of selection bias, which is measured by the Kullback-Leibler divergence. The larger the Kullback-Leibler divergence, the greater the distributional distance between treatment and control group and, thus, the larger the selection bias. The outcome is generated as $Y\mid X=\mathbf{x}, T=t \sim (\mathbf{w}^\top \mathbf{x} + t + n)$, where $\mathbf{w}\sim\mathcal U((-1,1)^{10\times 1})$, and $n\sim \mathcal N(0, 0.1)$.

\begin{figure}
	\includegraphics[scale=0.6]{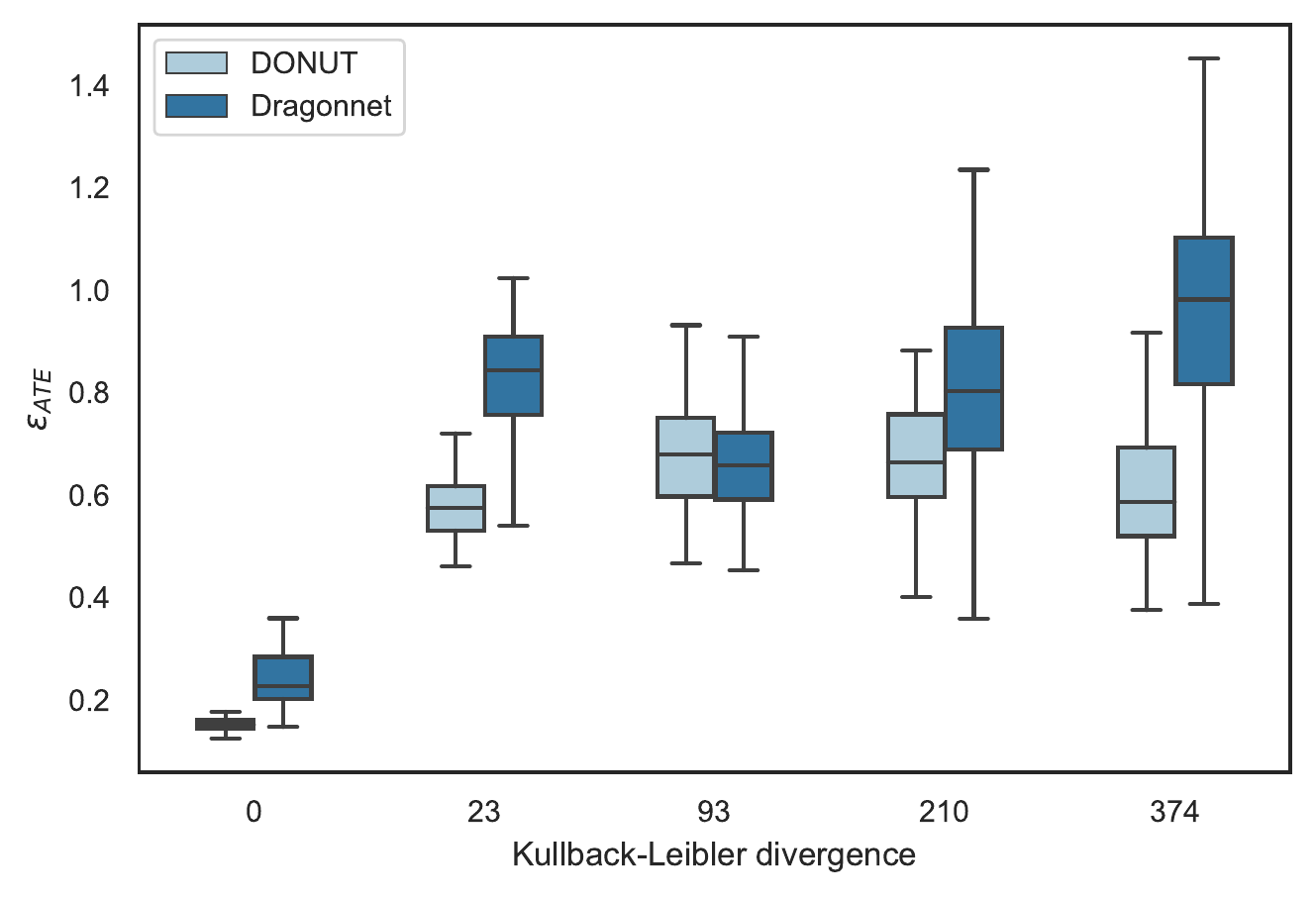}%
	\caption{Comparison of DONUT and Dragonnet in presence of selection bias (when treatment and control group differ substantially). Reported is the mean and standard deviation of $\epsilon_\text{ATE}$ on synthetic data as selection bias is varied (measured by the Kullback-Leibler divergence). For each level, both algorithms are run on 100 realizations of the datasets. Lower is better.}\label{fig:SimSelectionBias}%
\end{figure}

We compare DONUT against the current state-of-the-art neural network method, Dragonnet. Dragonnet makes use of the augmented inverse probability weighted estimator (AIPWE) to estimate the average treatment effect. Dragonnet is particularly appropriate for the comparison to DONUT, since it also builds upon the architecture of TARNet. For each level of selection bias, both algorithms are run on 100 realizations of the datasets. \Cref{fig:SimSelectionBias} presents the mean and standard deviation of $\epsilon_\text{ATE}$ for varying selection bias. We report two major insights: (i)~Dragonnet is outperformed by DONUT across different levels of selection bias. That is, when treatment and control groups differ substantially, DONUT achieves superior estimation of the ATE. Moreover, as selection bias increases, the estimation error of Dragonnet increases, whereas the estimation error of DONUT remains stable. (ii)~The standard deviation of both DONUT and Dragonnet increases as selection bias increases. However, the standard deviation of DONUT remains consistently smaller than the standard deviation of Dragonnet. In addition, this difference becomes more pronounced the larger the selection bias. This leads to a more reliable estimate of the ATE using DONUT than Dragonnet when treatment and control group differ substantially. As such, DONUT is more robust to selection bias.
\section{Conclusion}
Data-driven decision making is crucial in many domain such as epidemiology \cite{Robins2000a} or marketing \cite{hatt2020early}.  Previous methods estimate outcomes based on unconfoundedness but neglect any constraints that unconfoundedness imposes on the outcomes. In this paper, we present a novel regularization framework for estimating average treatment effects. For this, we exploit unconfoundedness during the estimation of the model parameters. Unconfoundedness requires the outcomes to be independent of the treatment assignment conditional on the covariates. We formalize this as an orthogonality constraint that requires the outcomes to be orthogonal to the treatment assignment. This orthogonality constraint is included in the loss function by perturbating the outcome functions.

Based on this, we develop DONUT, which leverages the predictive capabilities of neural networks to estimate average treatment effects. Experiments on a variety of benchmark datasets for causal inference demonstrate that DONUT outperforms the state-of-the-art substantially. This work provides interesting avenues for future research on estimating the ATE. We expect that most existing methods can be improved by incorporating orthogonal regularization. We leave the derivation of a unifying theory for future work.

\clearpage
\bibliography{library}

\begin{thebibliography}{51}
\providecommand{\natexlab}[1]{#1}
\providecommand{\url}[1]{\texttt{#1}}
\expandafter\ifx\csname urlstyle\endcsname\relax
  \providecommand{\doi}[1]{doi: #1}\else
  \providecommand{\doi}{doi: \begingroup \urlstyle{rm}\Url}\fi

\bibitem[Athey et~al.(2016)Athey, Imbens, and Wager]{athey2016approximate}
Athey, S., Imbens, G.~W., and Wager, S.
\newblock Approximate residual balancing: De-biased inference of average
  treatment effects in high dimensions.
\newblock \emph{arXiv preprint arXiv:1604.07125}, 2016.

\bibitem[Athey et~al.(2017)Athey, Imbens, Pham, and Wager]{athey2017estimating}
Athey, S., Imbens, G., Pham, T., and Wager, S.
\newblock Estimating average treatment effects: Supplementary analyses and
  remaining challenges.
\newblock \emph{American Economic Review}, 107\penalty0 (5):\penalty0 278--81,
  2017.

\bibitem[Austin(2011)]{austin2011introduction}
Austin, P.~C.
\newblock An introduction to propensity score methods for reducing the effects
  of confounding in observational studies.
\newblock \emph{Multivariate behavioral research}, 46\penalty0 (3):\penalty0
  399--424, 2011.

\bibitem[Belloni et~al.(2014)Belloni, Chernozhukov, and
  Hansen]{belloni2014inference}
Belloni, A., Chernozhukov, V., and Hansen, C.
\newblock Inference on treatment effects after selection among high-dimensional
  controls.
\newblock \emph{The Review of Economic Studies}, 81\penalty0 (2):\penalty0
  608--650, 2014.

\bibitem[Benkeser et~al.(2017)Benkeser, Carone, Laan, and
  Gilbert]{Benkeser2017}
Benkeser, D., Carone, M., Laan, M. J. v.~d., and Gilbert, P.~B.
\newblock {Doubly robust nonparametric inference on the average treatment
  effect}.
\newblock \emph{Biometrika}, 104\penalty0 (4):\penalty0 863--880, 2017.

\bibitem[Berrevoets et~al.(2020)Berrevoets, Jordon, Bica, Gimson, and van~der
  Schaar]{berrevoets2020organite}
Berrevoets, J., Jordon, J., Bica, I., Gimson, A., and van~der Schaar, M.
\newblock Organite: Optimal transplant donor organ offering using an individual
  treatment effect.
\newblock \emph{https://proceedings. neurips. cc/paper/2020}, 33, 2020.

\bibitem[Bertsimas et~al.(2017)Bertsimas, Kallus, Weinstein, and
  Zhuo]{bertsimas2017personalized}
Bertsimas, D., Kallus, N., Weinstein, A.~M., and Zhuo, Y.~D.
\newblock Personalized diabetes management using electronic medical records.
\newblock \emph{Diabetes care}, 40\penalty0 (2):\penalty0 210--217, 2017.

\bibitem[Breiman(2001)]{Breiman2001}
Breiman, L.
\newblock Random forests.
\newblock \emph{Machine learning}, 45\penalty0 (1):\penalty0 5--32, 2001.

\bibitem[Chernozhukov et~al.(2018)Chernozhukov, Chetverikov, Demirer, Duflo,
  Hansen, Newey, and Robins]{Chernozhukov2017a}
Chernozhukov, V., Chetverikov, D., Demirer, M., Duflo, E., Hansen, C., Newey,
  W., and Robins, J.
\newblock Double/debiased machine learning for treatment and structural
  parameters.
\newblock \emph{The Econometrics Journal}, 21\penalty0 (1):\penalty0 C1--C68,
  2018.

\bibitem[Chipman et~al.(2012)Chipman, George, and McCulloch]{Chipman2012a}
Chipman, H.~A., George, E.~I., and McCulloch, R.~E.
\newblock {BART: Bayesian additive regression trees}.
\newblock \emph{Annals of Applied Statistics}, 6\penalty0 (1):\penalty0
  266--298, 2012.

\bibitem[Chu et~al.(2020)Chu, Rathbun, and Li]{chu2020matching}
Chu, Z., Rathbun, S.~L., and Li, S.
\newblock Matching in selective and balanced representation space for treatment
  effects estimation.
\newblock In \emph{International Conference on Information \& Knowledge
  Management}, pp.\  205--214, 2020.

\bibitem[Crump et~al.(2008)Crump, Hotz, Imbens, and Mitnik]{Crump2008a}
Crump, R.~K., Hotz, V.~J., Imbens, G.~W., and Mitnik, O.~A.
\newblock {Nonparametric tests for treatment effect heterogeneity}.
\newblock \emph{Review of Economics and Statistics}, 90\penalty0 (3):\penalty0
  389--405, 2008.

\bibitem[Cui et~al.(2020)Cui, Shen, Li, Yao, Li, Chu, and Gao]{cui2020causal}
Cui, P., Shen, Z., Li, S., Yao, L., Li, Y., Chu, Z., and Gao, J.
\newblock Causal inference meets machine learning.
\newblock In \emph{SIGKDD International Conference on Knowledge Discovery \&
  Data Mining}, pp.\  3527--3528, 2020.

\bibitem[Farrell et~al.(2018)Farrell, Liang, and Misra]{Farrell2018}
Farrell, M.~H., Liang, T., and Misra, S.
\newblock {Deep neural networks for estimation and inference}.
\newblock \emph{arXiv preprint arXiv:1809.09953}, 2018.

\bibitem[Funk et~al.(2011)Funk, Westreich, Wiesen, St{\"{u}}rmer, Brookhart,
  and Davidian]{Funk2011}
Funk, M.~J., Westreich, D., Wiesen, C., St{\"{u}}rmer, T., Brookhart, M.~A.,
  and Davidian, M.
\newblock {Doubly robust estimation of causal effects}.
\newblock \emph{American Journal of Epidemiology}, 173\penalty0 (7):\penalty0
  761--767, 2011.

\bibitem[Hatt \& Feuerriegel(2020)Hatt and Feuerriegel]{hatt2020early}
Hatt, T. and Feuerriegel, S.
\newblock Early detection of user exits from clickstream data: A markov
  modulated marked point process model.
\newblock In \emph{Proceedings of The Web Conference 2020}, pp.\  1671--1681,
  2020.

\bibitem[Hatt \& Feuerriegel(2021)Hatt and Feuerriegel]{hatt2021sequential}
Hatt, T. and Feuerriegel, S.
\newblock Sequential deconfounding for causal inference with unobserved
  confounders.
\newblock \emph{arXiv preprint arXiv:2104.09323}, 2021.

\bibitem[Heckman et~al.(1997)Heckman, Ichimura, and Todd]{Heckman1997}
Heckman, J.~J., Ichimura, H., and Todd, P.~E.
\newblock {Matching as an econometric evaluation estimator: Evidence from
  evaluating a job training programme}.
\newblock \emph{The Review of Economic Studies}, 64\penalty0 (4):\penalty0
  605--654, 1997.

\bibitem[Hellewell et~al.(2020)Hellewell, Abbott, Gimma, Bosse, Jarvis,
  Russell, Munday, Kucharski, Edmunds, Sun, et~al.]{hellewell2020feasibility}
Hellewell, J., Abbott, S., Gimma, A., Bosse, N.~I., Jarvis, C.~I., Russell,
  T.~W., Munday, J.~D., Kucharski, A.~J., Edmunds, W.~J., Sun, F., et~al.
\newblock Feasibility of controlling covid-19 outbreaks by isolation of cases
  and contacts.
\newblock \emph{The Lancet Global Health}, 2020.

\bibitem[Imbens \& Rubin(2015)Imbens and Rubin]{imbens2015causal}
Imbens, G.~W. and Rubin, D.~B.
\newblock \emph{{Causal inference in statistics, social, and biomedical
  sciences}}.
\newblock Cambridge University Press, 2015.

\bibitem[Johansson et~al.(2016)Johansson, Shalit, and Sontag]{Johansson2016}
Johansson, F.~D., Shalit, U., and Sontag, D.
\newblock {Learning representations for counterfactual inference}.
\newblock In \emph{International Conference on Machine Learning}, pp.\
  4407--4418, 2016.

\bibitem[Kallus(2017)]{kallus2017recursive}
Kallus, N.
\newblock Recursive partitioning for personalization using observational data.
\newblock In \emph{International Conference on Machine Learning}, pp.\
  1789--1798, 2017.

\bibitem[Kallus(2018)]{kallus2018balanced}
Kallus, N.
\newblock Balanced policy evaluation and learning.
\newblock In \emph{Advances in Neural Information Processing Systems}, pp.\
  8895--8906, 2018.

\bibitem[Kang \& Schafer(2007)Kang and Schafer]{Kang2007}
Kang, J. D.~Y. and Schafer, J.~L.
\newblock {Demystifying double robustness: A comparison of alternative
  strategies for estimating a population mean from incomplete data}.
\newblock \emph{Statistical Science}, 22\penalty0 (4):\penalty0 523--539, 2007.

\bibitem[Kucharski et~al.(2020)Kucharski, Russell, Diamond, Liu, Edmunds, Funk,
  Eggo, Sun, Jit, Munday, et~al.]{kucharski2020early}
Kucharski, A.~J., Russell, T.~W., Diamond, C., Liu, Y., Edmunds, J., Funk, S.,
  Eggo, R.~M., Sun, F., Jit, M., Munday, J.~D., et~al.
\newblock Early dynamics of transmission and control of covid-19: a
  mathematical modelling study.
\newblock \emph{The lancet infectious diseases}, 2020.

\bibitem[Liu et~al.(2020)Liu, Yin, and Zhang]{liu2020estimating}
Liu, R., Yin, C., and Zhang, P.
\newblock Estimating individual treatment effects with time-varying
  confounders.
\newblock 2020.

\bibitem[Louizos et~al.(2017)Louizos, Shalit, Mooij, Sontag, Zemel, and
  Welling]{Louizos2017}
Louizos, C., Shalit, U., Mooij, J., Sontag, D., Zemel, R., and Welling, M.
\newblock {Causal effect inference with deep latent-variable models}.
\newblock In \emph{Advances in Neural Information Processing Systems}, pp.\
  6447--6457, 2017.

\bibitem[MacDorman \& Atkinso(1998)MacDorman and Atkinso]{macforman98}
MacDorman, M.~F. and Atkinso, J.~O.
\newblock {Infant mortality statistics from the linked birth/infant death}.
\newblock In \emph{Mon Vital Stat Rep, 46 (suppl 2)}, pp.\  1--22, 1998.

\bibitem[Mikosch et~al.(1997)Mikosch, der Vaart, and Wellner]{Mikosch1997}
Mikosch, T., der Vaart, A. W.~v., and Wellner, J.~A.
\newblock \emph{{Weak convergence and empirical processes}}, volume~92.
\newblock Springer Science {\&} Business Media, 1997.

\bibitem[Nie \& Wager(2017)Nie and Wager]{nie2017quasi}
Nie, X. and Wager, S.
\newblock Quasi-oracle estimation of heterogeneous treatment effects.
\newblock \emph{arXiv preprint arXiv:1712.04912}, 2017.

\bibitem[Pearl(2009)]{pearl2009causality}
Pearl, J.
\newblock \emph{Causality}.
\newblock Cambridge university press, 2009.

\bibitem[Prem et~al.(2020)Prem, Liu, Russell, Kucharski, Eggo, Davies, Flasche,
  Clifford, Pearson, Munday, et~al.]{prem2020effect}
Prem, K., Liu, Y., Russell, T.~W., Kucharski, A.~J., Eggo, R.~M., Davies, N.,
  Flasche, S., Clifford, S., Pearson, C.~A., Munday, J.~D., et~al.
\newblock The effect of control strategies to reduce social mixing on outcomes
  of the covid-19 epidemic in wuhan, china: a modelling study.
\newblock \emph{The Lancet Public Health}, 2020.

\bibitem[Rakesh et~al.(2018)Rakesh, Guo, Moraffah, Agarwal, and
  Liu]{rakesh2018linked}
Rakesh, V., Guo, R., Moraffah, R., Agarwal, N., and Liu, H.
\newblock Linked causal variational autoencoder for inferring paired spillover
  effects.
\newblock In \emph{International Conference on Information \& Knowledge
  Management}, pp.\  1679--1682, 2018.

\bibitem[Ray \& Szabo(2019)Ray and Szabo]{Ray2019a}
Ray, K. and Szabo, B.
\newblock Debiased bayesian inference for average treatment effects.
\newblock In \emph{Advances in Neural Information Processing Systems}, pp.\
  11929--11939, 2019.

\bibitem[Robins(1986)]{robins1986new}
Robins, J.
\newblock A new approach to causal inference in mortality studies with a
  sustained exposure period—application to control of the healthy worker
  survivor effect.
\newblock \emph{Mathematical modelling}, 7\penalty0 (9-12):\penalty0
  1393--1512, 1986.

\bibitem[Robins et~al.(2000)Robins, Hern{\'{a}}n, and Brumback]{Robins2000a}
Robins, J.~M., Hern{\'{a}}n, M.~A., and Brumback, B.
\newblock {Marginal structural models and causal inference in epidemiology}.
\newblock \emph{Epidemiology}, 11\penalty0 (5):\penalty0 550--560, 2000.

\bibitem[Rosenbaum \& Rubin(1983)Rosenbaum and Rubin]{Trust2016}
Rosenbaum, P.~R. and Rubin, D.~B.
\newblock The central role of the propensity score in observational studies for
  causal effects.
\newblock \emph{Biometrika}, 70\penalty0 (1):\penalty0 41--55, 1983.

\bibitem[Rubin(2005)]{Rubin2005}
Rubin, D.~B.
\newblock {Causal inference using potential outcomes: Design, modeling,
  decisions}.
\newblock \emph{Journal of the American Statistical Association}, 100\penalty0
  (469):\penalty0 322--331, 2005.

\bibitem[Shalit et~al.(2017)Shalit, Johansson, and Sontag]{Shalit2017a}
Shalit, U., Johansson, F.~D., and Sontag, D.
\newblock {Estimating individual treatment effect: Generalization bounds and
  algorithms}.
\newblock In \emph{International Conference on Machine Learning}, pp.\
  4709--4718, 2017.

\bibitem[Shi et~al.(2019)Shi, Blei, and Veitch]{Shi2019a}
Shi, C., Blei, D.~M., and Veitch, V.
\newblock {Adapting neural networks for the estimation of treatment effects}.
\newblock In \emph{Advances in Neural Information Processing Systems}, pp.\
  1--11, 2019.

\bibitem[Shimoni et~al.(2018)Shimoni, Yanover, Karavani, and
  Goldschmnidt]{Shimoni2018}
Shimoni, Y., Yanover, C., Karavani, E., and Goldschmnidt, Y.
\newblock Benchmarking framework for performance-evaluation of causal inference
  analysis.
\newblock \emph{arXiv preprint arXiv:1802.05046}, 2018.

\bibitem[Swaminathan \& Joachims(2015{\natexlab{a}})Swaminathan and
  Joachims]{swaminathan2015batch}
Swaminathan, A. and Joachims, T.
\newblock Batch learning from logged bandit feedback through counterfactual
  risk minimization.
\newblock \emph{The Journal of Machine Learning Research}, 16\penalty0
  (1):\penalty0 1731--1755, 2015{\natexlab{a}}.

\bibitem[Swaminathan \& Joachims(2015{\natexlab{b}})Swaminathan and
  Joachims]{swaminathan2015self}
Swaminathan, A. and Joachims, T.
\newblock The self-normalized estimator for counterfactual learning.
\newblock In \emph{Advances in Neural Information Processing Systems}, pp.\
  3231--3239, 2015{\natexlab{b}}.

\bibitem[van~der Laan \& Rose(2011)van~der Laan and Rose]{van2011targeted}
van~der Laan, M.~J. and Rose, S.
\newblock \emph{{Targeted learning: causal inference for observational and
  experimental data}}.
\newblock Springer Science \& Business Media, 2011.

\bibitem[van~der Laan \& Rubin(2006)van~der Laan and Rubin]{van2006targeted}
van~der Laan, M.~J. and Rubin, D.
\newblock Targeted maximum likelihood learning.
\newblock \emph{The international journal of biostatistics}, 2\penalty0 (1),
  2006.

\bibitem[Vansteelandt \& Joffe(2014)Vansteelandt and Joffe]{Vansteelandt2014a}
Vansteelandt, S. and Joffe, M.
\newblock {Structural nested models and G-estimation: The partially realized
  promise}.
\newblock \emph{Statistical Science}, 29\penalty0 (4):\penalty0 707--731, 2014.

\bibitem[Wager \& Athey(2018)Wager and Athey]{Wager2018a}
Wager, S. and Athey, S.
\newblock {Estimation and inference of heterogeneous treatment effects using
  random forests}.
\newblock \emph{Journal of the American Statistical Association}, 113\penalty0
  (523):\penalty0 1228--1242, 2018.

\bibitem[Wang et~al.(2017)Wang, Agarwal, and Dud{\i}k]{wang2017optimal}
Wang, Y.-X., Agarwal, A., and Dud{\i}k, M.
\newblock Optimal and adaptive off-policy evaluation in contextual bandits.
\newblock In \emph{International Conference on Machine Learning}, pp.\
  3589--3597. PMLR, 2017.

\bibitem[Yao et~al.(2018)Yao, Huai, Li, Gao, Li, and Zhang]{Yao2018a}
Yao, L., Huai, M., Li, S., Gao, J., Li, Y., and Zhang, A.
\newblock {Representation learning for treatment effect estimation from
  observational data}.
\newblock In \emph{Advances in Neural Information Processing Systems}, pp.\
  2633--2643, 2018.

\bibitem[Yoon et~al.(2018)Yoon, Jordon, and {van der Schaar}]{Yoon2018a}
Yoon, J., Jordon, J., and {van der Schaar}, M.
\newblock {Ganite: Estimation of individualized treatment effects using
  generative adversarial nets}.
\newblock In \emph{International Conference on Learning Representations}, pp.\
  1--22, 2018.

\bibitem[Zalinescu(2002)]{zalinescu2002convex}
Zalinescu, C.
\newblock \emph{Convex analysis in general vector spaces}.
\newblock World scientific, 2002.

\end{thebibliography}
\bibliographystyle{icml2021}

\newpage
\appendix
\onecolumn

\section{Results on ACIC 2018}\label{sec:acic_2018}
Similar to \citep{Shi2019a}, we further evaluate the performance of DONUT on ACIC 2018, which is a collection of semi-synthetic datasets derived from the linked birth and infant death data~(LBIDD) \citep{macforman98}, developed for the 2018 Atlantic Causal Inference Conference competition~(ACIC) \citep{Shimoni2018}. The simulation includes 63 different data generation processes with a sample size from 1,000 to 50,000. Each dataset is a realization from a separate distribution, which itself is randomly drawn in accordance with the settings of the data generation process. Similar to \citep{Shi2019a}, we randomly pick 3 datasets\footnote{We provide the list of unique dataset identification numbers at \url{github.com/tobhatt/donut} for reproducibility.} for each of the 63 data generating process settings of size either 5k or 10k and exclude all datasets with indication of strong selection bias. This yields a total number of 97 datasets. 

We average the results over 10 train/validation/test splits for each dataset for ACIC 2018, all with ratios 56/24/20. We use the absolute error in average treatment effect \citep{Shi2019a}: $\epsilon_\text{ATE} = \vert\frac{1}{n}\sm i n(f(x_i, 1) - f(x_i, 0)) - \frac{1}{n}\sm i n(\hat{f}(x_i, 1) - \hat{f}(x_i, 0))\vert$. We compare DONUT against the following 6 state-of-the-art methods for estimating treatment effects: Treatment-agnostic representation network~(\textbf{TARNet}) \citep{Shalit2017a}, \textbf{Dragonnet} \citep{Shi2019a}, double robust estimator~(\textbf{DRE}) \citep{athey2017estimating}, targeted maximum likelihood estimator~(\textbf{TMLE}) \citep{van2006targeted}, partially linear regression~(\textbf{PLR}) \citep{Chernozhukov2017a}, and double selection estimator~(\textbf{DSE}) \citep{athey2017estimating}. Where nuisance functions are needed, the outcome and propensity score model of Dragonnet is used. We do not compare to debiased Gaussian process~(\textbf{D-GP}) \citep{Ray2019a}, due to nonscalability of the author's code on large datasets. See Section 5 for results on IHDP, Twins, and Jobs datasets. \Cref{tbl:acic_results} presents the results of the experiments on ACIC 2018. The main observation is that DONUT achieves superior performance across a large collection of datasets. This result further confirms the benefit of including orthogonal regularization in the estimation procedure.
\begin{table}
	\caption{\label{tbl:acic_results} Results for estimating average treatment effects on ACIC 2018. Lower is better.}
	\begin{center}
		\begin{small}
			\begin{sc}
				\renewcommand{\arraystretch}{1.1}
				\begin{tabular}{lcc}
					\multicolumn{2}{l}{\bf{Results}} \\
					\toprule
					& \multicolumn{2}{c}{\textbf{ACIC} ($\epsilon_\text{ATE}$)} \\
					Method & In-s. & Out-s.\\
					\hline
					\textsc{TARNet} & $4.53 \pm 0.72$ & $4.48 \pm 0.74$\\	
					\textsc{DRE} & $2.79 \pm 1.45$ & $2.81 \pm 1.48$ \\
					\textsc{DSE} & $2.22 \pm 0.63$ & $2.22 \pm 0.62$\\
					\textsc{PLR} & $1.68 \pm 0.42$ & $1.76 \pm 0.40$ \\
					\textsc{TMLE} & $2.57 \pm 1.10$ & $3.25 \pm 1.42$ \\
					\textsc{Dragonnet} & $2.97 \pm 1.46$ & $2.99 \pm 1.48$\\
					\hline
					\addlinespace[0.75ex]
					\textbf{\textsc{DONUT}} & $\bf{1.22} \pm \bf{.33}$ & $\bf{1.26} \pm \bf{0.29}$\\					
					\bottomrule	
				\end{tabular}
			\end{sc}
		\end{small}
	\end{center}
	\vspace{-2em}
\end{table}

\section{Hyperparameter Sensitivity}\label{apx:sensitivity_lambda}
\begin{figure}
	\centering
	\includegraphics[scale=0.55]{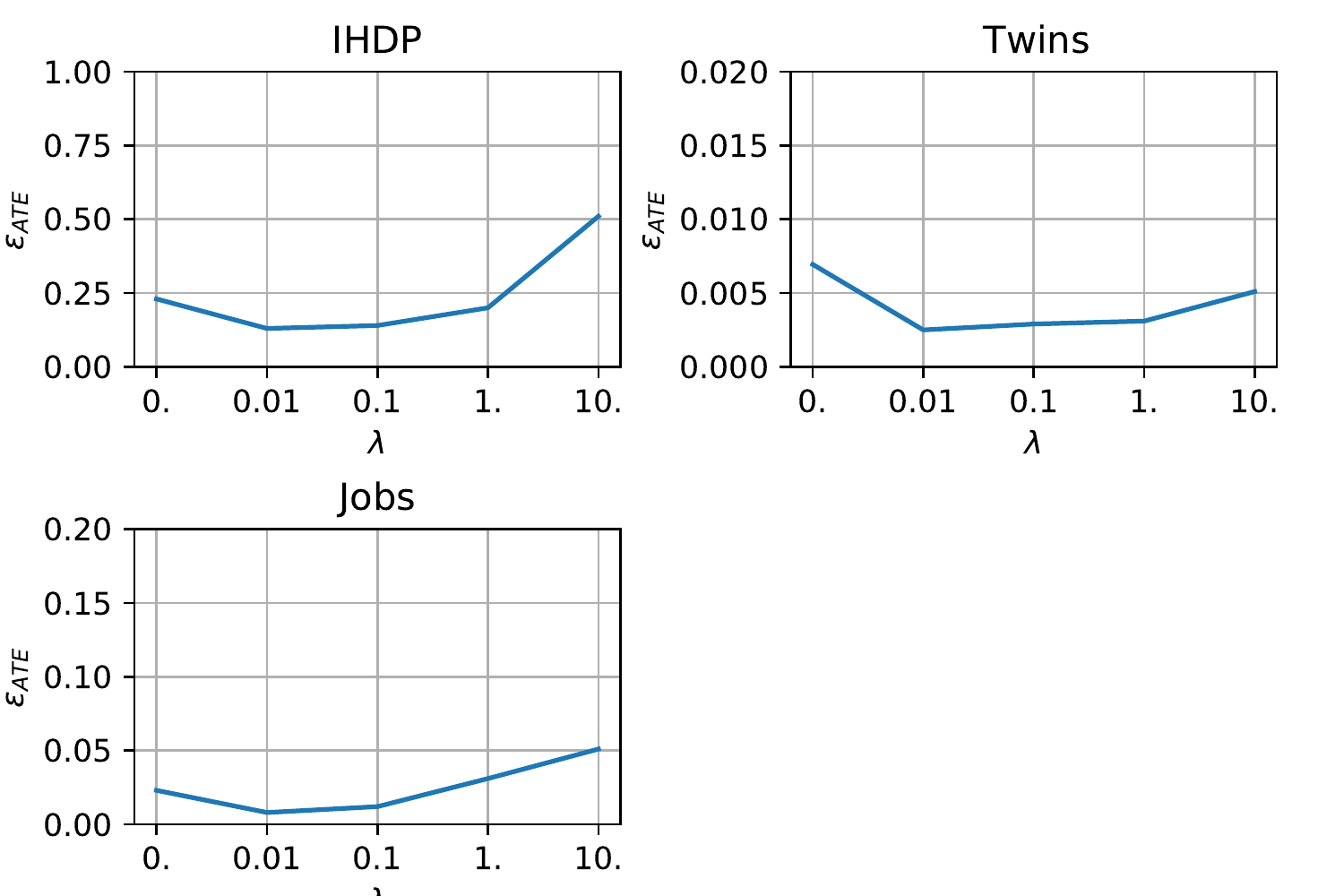}%
	\caption{Sensitivity analysis of the hyperparameter $\lambda$, which controls the strength of orthogonal regularization in the optimization problem (7). The in-sample error for each dataset IHDP, Twins, and Jobs is presented for $\lambda \in \{0, 0.01, 0.1, 1, 10\}$. Lower is better.}\label{fig:LambdaStudy}%
\end{figure}

We conduct a sensitivity analysis for the hyperparameter $\lambda$, which controls the strength of orthogonal regularization in the optimization problem (7). For this, we compare in-sample errors of DONUT for different values of $\lambda$ on each of the datasets IHDP, Twins, and Jobs. We compare the following values for $\lambda \in \{0, 0.01, 0.1, 1, 10\}$, where $\lambda=0$ corresponds to DONUT without orthogonal regularization as presented in the ablation study in Section 5.3. We present the results in \Cref{fig:LambdaStudy} and observe that, expect for $\lambda=0$ and large values of $\lambda$, the values for DONUT for different values of $\lambda$ remains stable.

\section{Sufficiency of Regularizing the Untreated Outcome}\label{apx:OnlyTreatmentfree}
It is straightforward to adapt the orthogonal regularizer in \labelcref{eq:regularizer} to accommodate both outcomes. Consider the following adapted orthogonal regularizer
\begin{equation}\label{eq:AdaptedRegularizer}
	\Omega_{\mathit{OR}}(f, \pi, \epsilon; \dat) = \frac{1}{n}\sm i n (Y_i^\ast(0) - f^{\epsilon}(0, X_i))^2 + \frac{1}{n}\sm i n (Y_i^\ast(1) - f^{\epsilon}(1, X_i))^2,
\end{equation}
with
\begin{equation}\label{eq:AdaptedATEproxy2}
	Y_i^\ast(0) = Y_i - \psi^\ast\,T_i, \, Y_i^\ast(1) = Y_i + \psi^\ast\,(1-T_i), \, \psi^\ast = \frac1n \sm i n{f(X_i, 1) - f(X_i, 0)}
\end{equation}
and 
\begin{equation}\label{eq:AdaptedOrthogonality_enforce1}
	f^{\epsilon}(X_i, 0) = f(X_i, 0) + \epsilon(T_i - \pi(X_i)),
\end{equation}
\begin{equation}\label{eq:AdaptedOrthogonality_enforce2}
	f^{\epsilon}(X_i, 1) = f(X_i, 1) + \epsilon(T_i - \pi(x_i)),
\end{equation}
where $\epsilon$ is a additional model parameters. Similar to the result in \Cref{sec::proposed_framework}, this yields that $\langle Y^\ast(0) - f^{\epsilon}(X, 0), T - \pi(X) \rangle$ and $\langle Y^\ast(1) - f^{\epsilon}(X, 1), T - \pi(X) \rangle$ are zero, when the partial derivative of \labelcref{eq:AdaptedRegularizer} \wrt $\epsilon$ is zero.

We show that it is sufficient to regularize $Y(0)$, since if we can ensure that the inner product in \labelcref{eq:orthogonality} of $Y(0)$ and $T$ is zero, then it is also zero between $Y(1)$ and $T$. To this end, suppose the inner product of $Y(0)$ and $T$ is zero, \ie,  
\begin{equation}\setcounter{equation}{15}
	\langle Y(0) - f(X, 0), T - \pi(X)\rangle = 0.
\end{equation}

Then, using that $Y(1) = Y + \psi\,(1-T)$,
\begin{align}
	&\langle Y(1) - f(X, 1), T - \pi(X)\rangle\\
	&=\frac1n \sm i n (Y_i(1) - f(X_i, 1))(T_i - \pi(X_i))\\
	&=\frac1n \sm i n (Y_i-\psi_iT_i - f(X_i, 0) + \psi_i - f(X_i, 1) + f(X_i, 0))(T_i - \pi(X_i))\\
	&=\frac1n \sm i n Y_i(0) - f(X_i, 0)))(T_i - \pi(X_i))=0.
\end{align}
Hence, the inner product in \labelcref{eq:orthogonality} of $Y(1)$ and $T$ is zero as well.

%


\section{Proof of Proposition 1}\label{apx:proof:AsymptoticNormality}

First, we need to introduce some notation. Throughout the proof, we will use $\mathbb P\{f(Z)\} = \int f(z)d\mathbb P$ to denote expectations of $f(Z)$ for a random variable $Z$ (treating the function $f$ as fixed). Hence, $\mathbb P\{\hat f(Z)\}$ is random if $\hat f$ is random (\eg, estimated from the sample). In contrast, $\E[\hat f(Z)]$ is a fixed non-random quantity, which averages over randomness in both $Z$ and $\hat f$ and thus will not equal $\mathbb P\{\hat f(Z)\}$ except when $\hat f = f$ is fixed and non-random. Moreover, we let $\mathbb P_n = \frac{1}{n} \sm i n \delta_{Z_i}$ denote the empirical measure such that sample averages can be written as $\frac{1}{n}\sm i n f(Z_i) = \int f(z) d\mathbb P_n = \empPrb{f(Z)}$. For clarity, we will denote the true nuisance functions and average treatment effect as $\eta_0 = (f_0, \pi_0)$ and $\psi_0$, respectively. We further denote the triplet $(Y, T, X)$ by $Z$.

For $\hat f=f^\epsilon$, $\hat \pi$, and $\hat \psi = \frac{1}{n} \sm i n f^\epsilon(X_i, 1) - f^\epsilon(X_i, 0)$ from (9), the regularization framework yields
\begin{equation}
	\langle Y^\ast(0) - \hat f(X, 0), T - \hat\pi(X)\rangle = \frac{1}{n}\sm i n (Y_i -\hat\psi\,T_i - \hat f(X_i, 0))(T_i - \hat\pi(X_i)) = 0,\nonumber
\end{equation} 
by construction. Hence, we receive an analytical expression of the estimator of the average treatment effect by solving the above for $\hat\psi$, \ie, 
\begin{equation}\label{apx:eq:anal_expression}
	\hat\psi = \frac{\frac{1}{n}\sm i n (Y_i - \hat f(X_i, 0))(T_i - \hat\pi(X_i))}{\frac{1}{n}\sm i n T_i(T_i-\hat\pi(X_i))}
\end{equation} 
Therefore, the estimator is given by $\hat\psi = \empPrb{m(Z; \hat{\eta})}$, where
\begin{equation}
	m(Z; \eta) = \frac{(Y - f(X, 0))(T-\pi(X))}{\mathbb P\{T(T-\pi(X))\}},
\end{equation}
and $\eta=(\pi, f)$ denotes the nuisance functions. Because of Condition 1 in the roposition, at least one nuisance estimator needs to converge to the correct function, but one can be misspecified. Then, $\mathbb P\{m(Z; \bar{\eta})\} = \mathbb P\{m(Z; \eta_0)\} = \psi_0$, from the straightforward to check fact that $\mathbb P\{(m(Z; \bar{\pi}, f_0)\} = \mathbb P\{m(Z; \pi_0, \bar{f})\}$ for any $\bar{\pi}$ and $\bar{f}$. Consider the decomposition
\begin{align*}\label{eq:asympComp}
	\hat\psi - \psi_0 &= \empPrb{m(Z;\hat{\eta})} - \mathbb P\{ m(Z;\bar\eta)\}\\
	&=(\mathbb P_n-\mathbb{P})\{m(Z;\hat{\eta})\} + \mathbb P\{m(Z;\hat{\eta}) - m(Z;\bar\eta)\}.
\end{align*}
If the estimators for the nuisance functions $\hat{\eta}$ take values in $\mathbb P$-Donsker classes, then $m(Z;\hat{\eta})$ also belongs to a $\mathbb P$-Donsker class, since Lipschitz transformations of Donsker functions are again Donsker functions. The Donsker property together with the continuous mapping theorem yields that $(\mathbb P_n-\mathbb{P})\{m(Z;\hat{\eta})\}$ is asymptotically equivalent to $(\mathbb P_n-\mathbb{P})\{m(Z;\bar{\eta})\}$ up to $o_p(1/\sqrt{n})$\footnote{$o_p(1/r_n)$ employs the usual stochastic order notation so that $X_n=o_p(1/r_n)$ means that $r_nX_n \rightarrow 0$ in probability.} error (see \citep{Mikosch1997} for more details on $\mathbb P$-Donsker classes). Therefore,
\begin{equation}
	\hat\psi - \psi_0 = (\mathbb P_n-\mathbb{P})\{m(Z;\bar{\eta})\} + \mathbb P\{m(Z;\hat{\eta}) - m(Z;\bar\eta)\} + o_p(1/\sqrt{n}).
\end{equation}
It is left to show that $\mathbb P\{m(Z;\hat{\eta}) - m(Z;\bar\eta)\}$ is asymptotically negligible. This term equals
\begin{align}
	&\mathbb P\left\{\frac{(T-\hat\pi(X))(Y - \hat f(X,0))}{\mathbb P\{T(T-\hat\pi(X))\}} - \frac{(T-\pi_0(X))(Y - f_0(X,0))}{\mathbb P\{T(T-\pi_0(X))\}}
	\right\}\\
	&=\mathbb P\left\{\pi_0(X)\frac{(1-\hat\pi(X))(f_0(X,1) - \hat f(X,0))}{\mathbb P\{T(T-\hat\pi(X))\}}\right\}\\
	& - \mathbb P\left\{\pi_0(X)\frac{(1-\pi_0(X))(f_0(X,1) - f_0(X,0))}{\mathbb P\{T(T-\pi_0(X))\}}\right\}\nonumber\\
	&-\mathbb P\left\{\frac{(1-\pi_0(X))\hat\pi(X)(f_0(X,0) - \hat f(X,0))}{\mathbb P\{T(T-\hat\pi(X))\}}\right\}
\end{align}
\begin{align}
	&=\mathbb P\left\{\pi_0(X)\frac{(1-\hat\pi(X))(f_0(X,1) - \hat f(X,0))}{\mathbb P\{\pi_0(X)(1-\hat\pi(X))\}}\right\}\\ 
	&- \mathbb P\left\{\pi_0(X)\frac{(1-\pi_0(X))(f_0(X,1) - f_0(X,0))}{\mathbb P\{\pi_0(X)(1-\pi_0(X))\}}\right\}\nonumber\\
	&-\mathbb P\left\{\frac{(1-\pi_0(X))\hat\pi(X)(f_0(X,0) - \hat f(X,0))}{\mathbb P\{\pi_0(X)(1-\hat\pi(X))\}}\right\}\\
	&=\mathbb P\left\{\frac{\pi_0(X)(1-\hat\pi(X))(f_0(X,1) - \hat f(X,0))}{\mathbb P\{\pi_0(X)(1-\hat\pi(X))\}}\right\} - \psi_0\\
	&-\mathbb P\left\{\frac{(1-\pi_0(X))\hat\pi(X)(f_0(X,0) - \hat f(X,0))}{\mathbb P\{\pi_0(X)(1-\hat\pi(X))\}}\right\}\\
	&=\mathbb P\left\{\frac{\pi_0(X)(1-\hat\pi(X))(f_0(X,1) - \hat f(X,0))}{\mathbb P\{\pi_0(X)(1-\hat\pi(X))\}}\right\}\\
	&- \mathbb P\left\{\frac{\pi_0(X)(1-\hat\pi(X))(f_0(X,1) - f_0(X,0))}{\mathbb P\{\pi_0(X)(1-\hat\pi(X))\}}\right\}\nonumber\\
	&-\mathbb P\left\{\frac{(1-\pi_0(X))\hat\pi(X)(f_0(X,0) - \hat f(X,0))}{\mathbb P\{\pi_0(X)(1-\hat\pi(X))\}}\right\},
\end{align}where we use $\mathbb P\{m(Z; \bar{\eta})\} = \mathbb P\{m(Z; \eta_0)\}$, iterated expectation, and $\mathbb P\{T(T-\hat\pi(X))\} = \mathbb P\{\pi_0(X)(1-\hat\pi(X))\}$ (likewise for $\pi_0$ instead of $\hat\pi$). We further use that $\psi_0(X) = \psi_0$ and, therefore, $f_0(X,1) - f_0(X,0) = \psi_0$. As a result, by simplifying, the above equals
\begin{equation}
	\frac{\pi_0(X) - \hat\pi(X)}{\mathbb P\{\pi_0(X)(1-\hat \pi(X))\}} (f_0(X, 0) - \hat f(X, 0)).
\end{equation}
Therefore, by the fact that $\pi_0$ and $\hat\pi$ are bounded away from zero and one, along with the Cauchy-Schwarz inequality (\ie, $\mathbb P\{fg\}\leq\| f\|\| g\|$), we have that (up to a multiplicative constant) $\lvert \mathbb P \{m(Z;\hat \eta) - m(Z;\bar\eta)\}\rvert$ is bounded above by
\begin{equation}
	\|\pi_0(X) - \hat\pi(X)\| \| f_0(X, 0) - \hat f(X, 0)\|.
\end{equation}
Since $\lVert \hat{\eta} - \bar{\eta} \rVert = o_p(1/\sqrt{n})$, the product term $\mathbb P\{m(Z;\hat \eta) - m(Z;\bar \eta)\} = o_p(1/\sqrt{n})$ is asymptotically negligible. Then the estimator satisfies $\hat\psi - \psi_0 = (\mathbb P_n - \mathbb P)\{m(Z;\eta_0)\} + o_p(1/\sqrt{n})$. Thus, this proves the convergence of $\hat\psi$. It remains to show that the asymptotic variance of $\hat \psi$ equals $\sigma^2/\mathbb P\{\text{Var}(T\mid X)\}.$ Note that in the main paper, $\Efb{f(Z)}$ is used similarly to $\mathbb P\{f(Z)\}$. We introduced the notion $\mathbb P\{f(Z)\}$ here to make clear that we only integrate over randomness in $Z$ and not the function estimates $\hat f$. Using that $\mathbb P\{m(Z;\eta) - \psi\} = 0$ (by definition) yields
\begin{align}
	\text{Var}(\phi(Z;\psi,\eta)) = \mathbb P\{\phi(Z;\psi,\eta)^2\}
	&=\mathbb P\{(m(Z;\eta) - \psi)^2\}\\
	&=\mathbb P\{(m(Z;\eta)^2\} - \psi^2.
\end{align}
This, together with
\begin{align}
	\mathbb P\{m(Z;\eta)^2\} = \frac{\mathbb P\{U^2\}}{\mathbb P\{\text{Var}(T\mid X)\}} + \psi^2,
\end{align}
where $U=Y-f(X, T) \sim \mathcal N(0, \sigma^2)$ and $\mathbb P\{T(T-\pi(X))\} = \mathbb P\{\text{Var}(T\mid X)\}$ proves the statement.\qed

\end{document}